# Recovery of Images with Missing Pixels using a Gradient Compressive Sensing Algorithm

Isidora Stanković, *Student Member IET*

**Abstract**: This paper investigates the possibility of reconstruction of images considering that they are sparse in the DCT transformation domain. Two approaches are considered. One when the image is pre-processed in the DCT domain, using 8x8 blocks. The image is made sparse by setting the smallest DCT coefficients to zero. In the other case the original image is considered without pre-processing, assuming the sparsity as intrinsic property of the analyzed image. A gradient based algorithm is used to recover a large number of missing pixels in the image. The case of a salt-and-paper noise affecting a large number of pixels is easily reduced to the case of missing pixels and considered within the same framework. The reconstruction of images affected with salt-and-paper impulsive is compared with the images filtered using a median filter. The same algorithm can be used considering transformation of the whole image. Reconstructions of black and white and colour images are considered.

## 1. Introduction

One of the basic signal processing and information theory results is the sampling or Shannon-Nyquist theorem. This theorem has provided a possibility to represent a continuous-time signal by a set of numbers representing the signal values at discrete time instants. The Nyquist theorem states that as long as the sampling frequency is at least twice as high as the maximum frequency present in the signal the signal can be fully recovered from the samples. If that requirement is met, the reconstruction will be alias-free. Shannon-Nyquist is a widely used technique in digital signal processing. The sampling theorem can be restated in the following way. If the Fourier transform of a signal is not of an infinite duration, but it is located within a frequency range defined by a maximal absolute negative value and maximal positive value, then in order to reconstruct the signal we do not need to know all signals values. It is sufficient to know just the signal values at some time instances defined by the sampling theorem.

Compressive Sensing (or Compressed Sampling) is a new, growing field in signal processing with an intensive development in the last ten years. In the centre of compressive sensing is the notion of a sparse signal. As it is known a signal can be transformed to different domains with a possibility to obtain the exact signal via the inverse transform. A signal is sparse in a transformation domain if the number of non-zero coefficients in that domain is much fewer than the number of signal samples. For a signal that has just a small number of non-zero coefficients in a transformation domain we can expect that a small number of non-zero coefficients can be recovered from a small set of signal samples. This is exactly the basic goal of the CS, to use a small number of randomly positioned samples to reconstruct all values of a sparse signal [1]-[6]. Since the introduction of CS, various techniques are developed to reconstruct a signal. They belong to two large groups. One of them is based on the signal analysis and reconstruction in the transformation domain where the signal is sparse and the other one is gradient based. The CS is already used in many signal processing areas. Image-related applications, such as medical imaging, signal compression, denoising, photography, holography, facial recognition, radar and array signal processing are some of the applications.

Real images usually have just a few DCT coefficients within each transformation block that should be considered as non-zero. Thus, they comply with the CS algorithm requirement that the image is a sparse signal in the considered transformation domain [7]. An approach of reconstructing when a large number of the missing pixels is randomly positioned over the whole image will be considered.

The paper is organized as follows. The problem formulation is given in the section. A reconstruction algorithm, with image recovery examples, is presented in Section 3. The case of impulsive salt-and-paper noise is redefined into the CS framework in Section 4. The results are compared with median filtered results as well. The conclusions are given in Section 5.

## 2. Basic Theory

The approach considered in this paper deals with an image whose random set of pixels is available. The task is to reconstruct the pixels that are not available. To make the reconstruction possible, the picture sparsity in a transformation (in this case DCT) domain should be assumed. The DCT of an image is commonly calculated using its into 8x8 blocks. Most of the images could be considered as sparse in the DCT domain without any processing. If we want to be sure that the original image, which will be processed in our examples, is sparse we can calculate the DCT of its 8x8 blocks and make some of the coefficients, with lowest amplitudes, to be equal to zero. Making the image sparse in the DCT domain in this way will not make a big visual difference with respect to the original image. Note that even if we do not use this step, most of the images are sparse in the DCT domain, and the presented methods can be used without pre-processing of the original image. In Figure 1 the original image "Isi" and a version with pre-processed DCT coefficients to make it exactly sparse are presented. These images will be used for most of our analysis and reconstruction.

An image that is sparse in a transformation domain can be reconstructed from a very reduced set of available pixels. Assume that the image values are available only for the pixels

$$x(n,m), \text{ for } (n,m) = (n_1,m_1),(n_2,m_2),...,(n_M,m_M) \quad (1)$$

Based on the available pixels we form an image $y(n,m)$ that assumes the values of the original image at the positions of available pixels. For the pixels that are not available we will set their value to zero. This new image is:

$$y(n,m) = \begin{cases} x(n,m), & \text{if } (n,m) = (n_1,m_1),...,(n_M,m_M) \\ 0, & \text{elsewhere} \end{cases} \quad (2)$$

For missing pixels we can assume any value in the initial step, since the algorithm should reconstruct the true image values at these positions. For graphical representation of missing pixels we will use the value 255 instead of 0. The missing pixels are then represented by the white colour (blank), as in Figure 2.

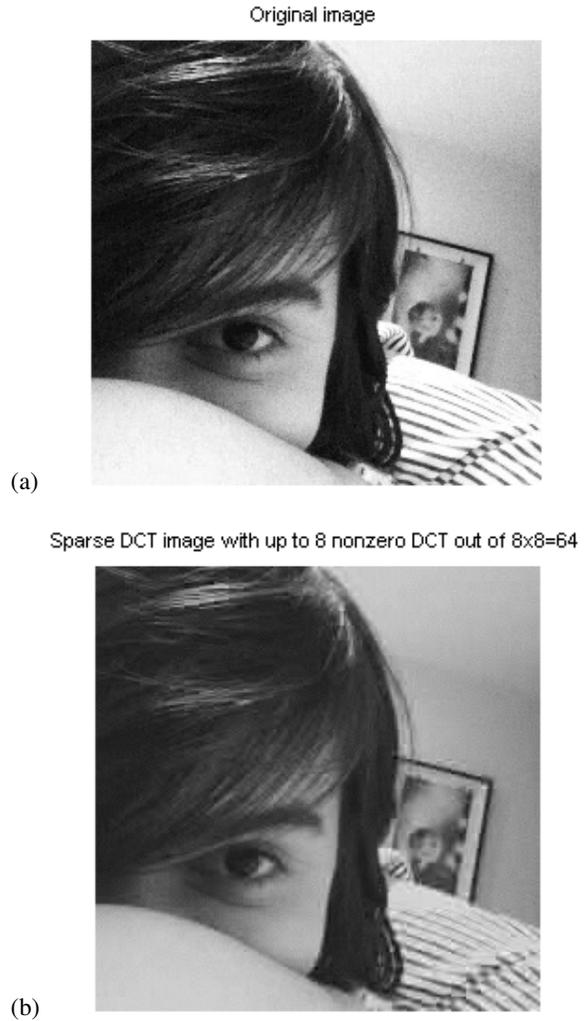

Figure 1: (a) Original image; (b) Image sparse in the DCT domain

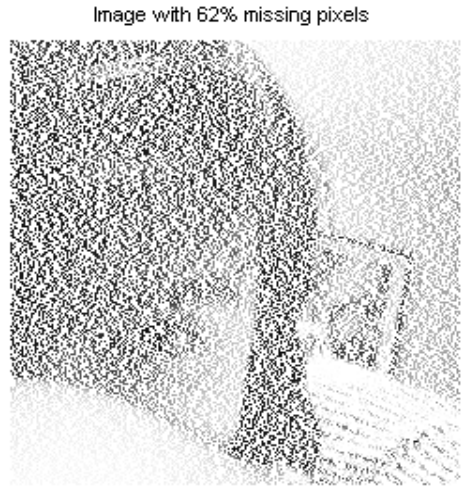

Figure 2: Image with a large number of missing pixels

## 3. Reconstruction Algorithm

The gradient algorithm is used on 8x8 blocks of the picture. The blocks with reconstructed pixels are then recombined into the final image. The two-dimensional gradient algorithm is similar to the one-dimensional one [8]-[10] except that we use the position of the value in x (rows) **and** y (columns) directions rather than a value in a sequence. The $L_1$-norm is used over two-dimensional space of image, i.e. as a sum the DCT values over the rows and columns.

Algorithm is implemented as follows [11]:

- For the available image pixels $x(n,m)$, for $n = n_1, n_2, ..., n_M$ and $m = m_1, m_2, ..., m_M$, a signal $y(n,m) = x(n,m)$ is formed, within one block, where $n$ represents the row and $m$ represents column of the image pixel. Other values are set to zero:

$$y(n,m) = \begin{cases} x(n,m), & \text{if } (n,m) = (n_1,m_1),...,(n_M,m_M) \\ 0, & \text{elsewhere} \end{cases} \quad (3)$$

- Then for each of $k \neq n_1, n_2, ..., n_M$ and each of $l \neq m_1, m_2, ..., m_M$ within the same block we form:

$$y_1^{(k,l)}(n,m) = y(n,m) + \Delta\delta(n-k, m-l)$$
$$y_2^{(k,l)}(n,m) = y(n,m) - \Delta\delta(n-k, m-l) \quad (4)$$

- The 2D-DCT is calculated as

$$Y_1^{(k,l)}(p,q) = \|DCT2\{y_1^{(k,l)}(n,m)\}\|_1$$
$$Y_2^{(k,l)}(p,q) = \|DCT2\{y_2^{(k,l)}(n,m)\}\|_1 \quad (5)$$

Where $L_1$-norm of 2D-DCT is:

$$\|Y_1(k,l)\|_1 = \sum_{k=0}^{N-1}\sum_{l=0}^{M-1} |Y_1^{(k,l)}(p,q)|$$
$$\|Y_2(k,l)\|_1 = \sum_{k=0}^{N-1}\sum_{l=0}^{M-1} |Y_2^{(k,l)}(p,q)| \quad (6)$$

The gradient corresponding to the change of the $(k,l)$-th pixel $k \neq n_1, n_2, ..., n_M$ and $l \neq m_1, m_2, ..., m_M$ is:

$$Gr(k,l) = \frac{\|Y_1(k,l)\|_1 - \|Y_2(k,l)\|_1}{2\Delta} \quad (7)$$

Each missing signal value $k \neq n_1, n_2, ..., n_M$, $l \neq m_1, m_2, ..., m_M$ is then changed in the direction of the gradient for a step $\mu$

$$y^{(i+1)}(k,l) = y^{(i)}(k,l) - \mu Gr(k,l) \quad (8)$$

The available signal values are not changed.

When the given parameters $\Delta, \mu$ do not improve the result, the step size is reduced. For example, as $\Delta \rightarrow \Delta/10$ and $\mu \rightarrow \mu/10$. The procedure is repeated until the desired reconstruction accuracy is achieved.

Applying the algorithm on the sparse image, the reconstructed image is obtained, Figure 3.

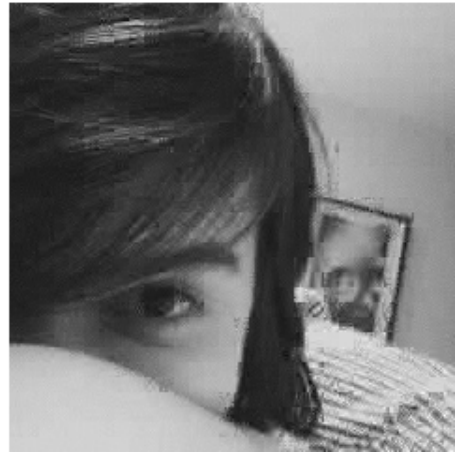

Figure 3: Reconstructed image using a CS gradient algorithm

The algorithm was tested on the same image but with 75% of missing samples. The image with 75% missing samples is presented in Figure 4. The picture already became blurrier. The 8x8 blocks started to be more visible. We can conclude that the algorithm is working properly with up to 75% missing samples, but not more.

## 4. Filtering Impulsive Noise

In some applications the signal is disturbed by a strong impulsive noise in images known as "salt and pepper". The image affected by a large number of impulsive noise pixels has many black and white pixels, manifesting themselves as salt and pepper noise. A classical approach to reconstruct this picture is in using a Median filter.

To illustrate how Median filter works, image "Isi" is used shown in Figure 6 (a). The image affected with 100 salt and pepper pixels (out of 240x240= 57600 pixels), shown in Figure 6 (b), is considered.

Doing the same procedure as for 75% missing samples, the reconstructed image is shown in Figure 5.

Using both 3x3 and 5x5 Median filter, the picture is filtered and shown in Figure 7.

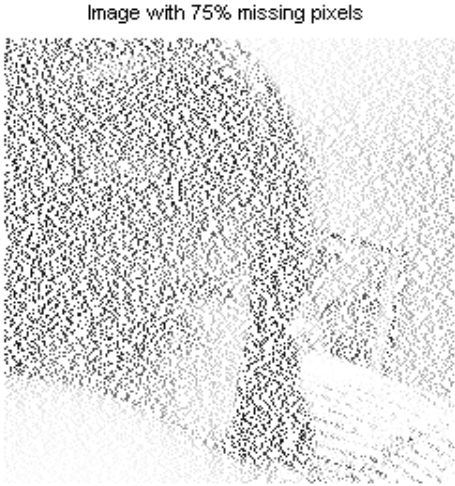

Figure 4: Image with 75% missing samples

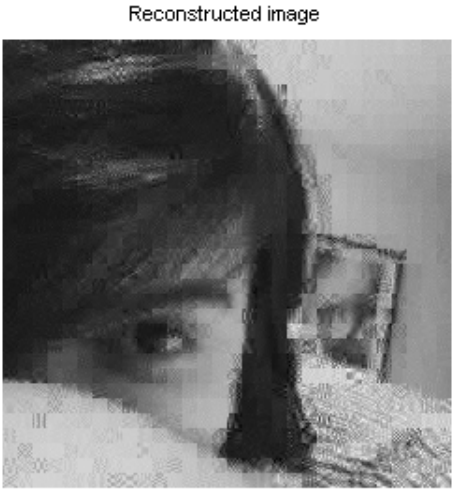

Figure 5: Reconstructed image from the image with 75% missing samples

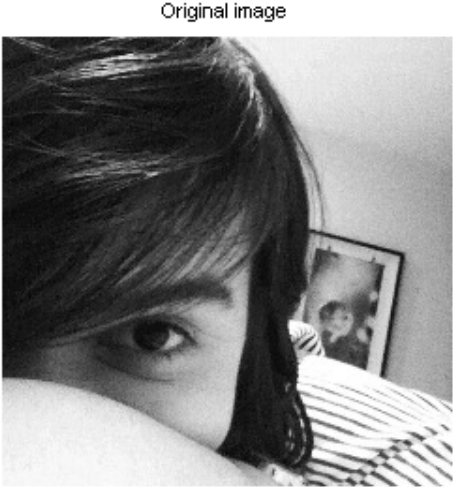

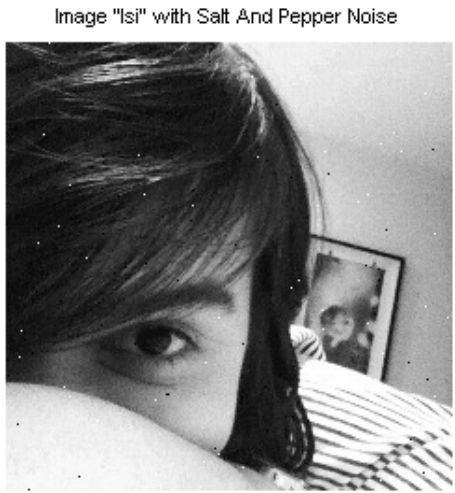

Figure 6: (a) Original image; (b) Image affected with a salt-and-paper noise in small number of pixels

Both (a) and (b) removed the salt and pepper noise but the image became more blurry in both cases. If the picture is highly affected with noise, as shown in Figure 8 the Median filter will not give as good results as before. The reconstruction of the image using Median filter is shown in Figure 9

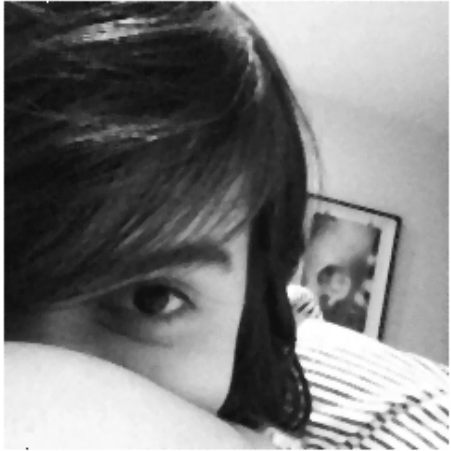

(a)

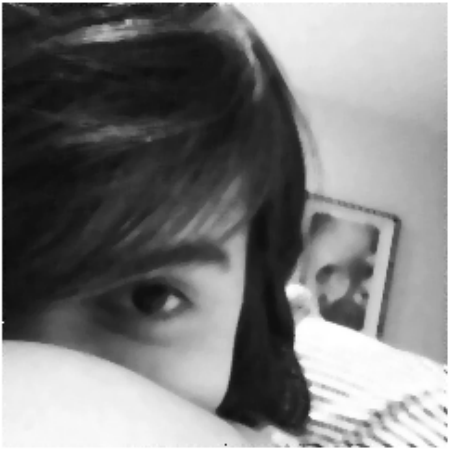

(b)

Figure 7: Filtered image using (a) 3x3 median filter; (b) 5x5 median filter

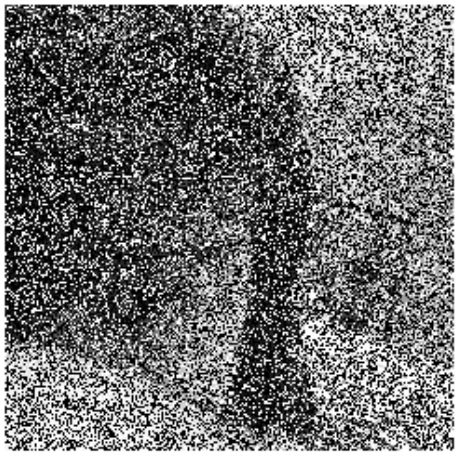

Figure 8: Image with salt and pepper noise in a large number of pixels

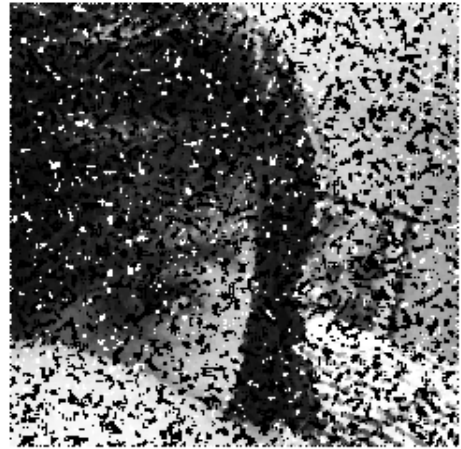

(a)

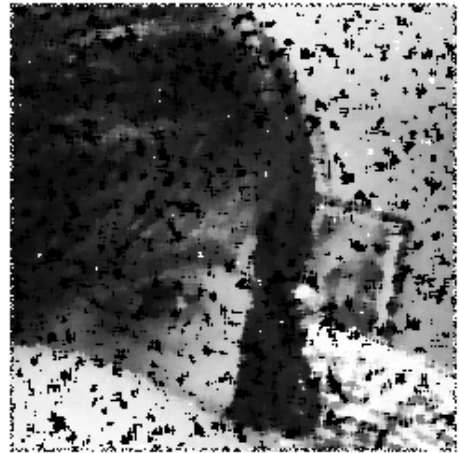

(b)

Figure 9: Filtered image using (a) 3x3 median filter; (b) 5x5 median filter

We can conclude that both 3x3 and 5x5 Median filters not only let some noise to pass, but also blurred the picture, degrading quality. Next we will show that these pictures can be reconstructed using the CS algorithms introduced before.

In Compressive Sensing, we define the noisy pixels of the image as missing and do the reconstruction with the algorithm defined earlier [12]-[15]. Let us consider the same noisy image as in Figure 8. We know that the salt and pepper noise is either black or white, so in pixel values it is either 0 or 255 respectively. The assumption made is then:

if $y(n,m) = 255$, signal sample discarded,
initial value set to neutral value $y(n) = 128$

(9)

if $y(n,m) = 0$, signal sample discarded,
initial value set to $y(n) = 128$

In this case, for missing samples the value 255 is used instead of 0. It represents the white colour (blank).

The image with impulsive noise and the image with discarded samples are shown in Figure 10.

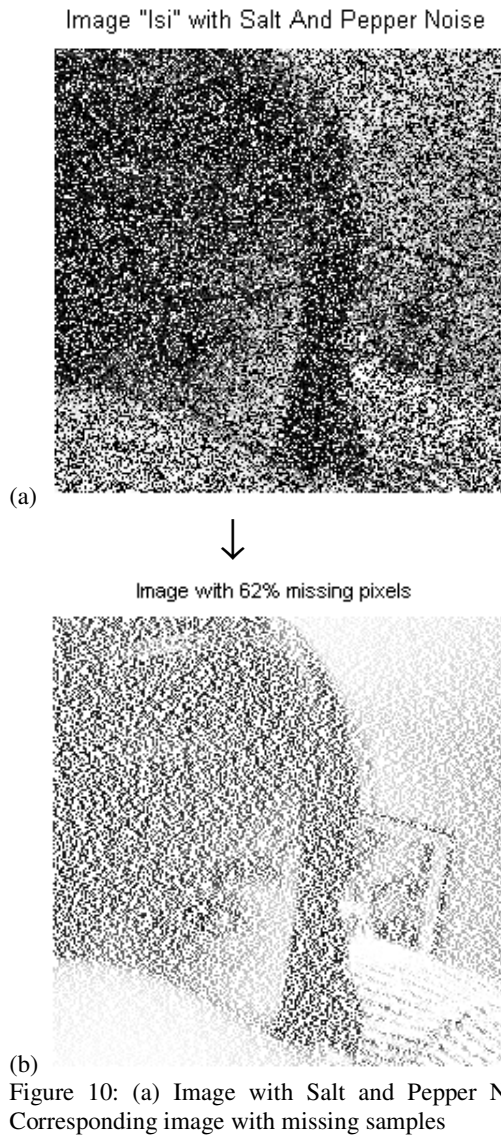

(a)

↓

(b)

Figure 10: (a) Image with Salt and Pepper Noise; (b) Corresponding image with missing samples

After going through the algorithm from equations (3)-(8) the new, reconstructed image is shown in Figure 11.

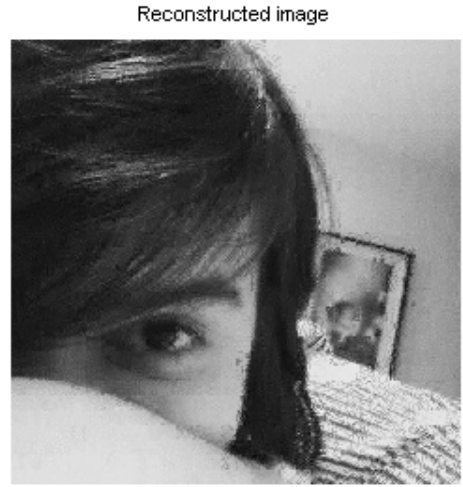

Figure 11: Reconstructed image using a CS gradient algorithm

The algorithm is also tested on colour images. For a colour images, each primary colour (red, green and blue) is reconstructed separately. The presented reconstruction method is used without pre-processing of the original image to make it sparse in an artificial manner by setting the DCT coefficients to zero. The number of corrupted pixels is 50% of the total number of pixels. The reconstruction of colour image "Autumn" is shown in Figure 12.

## 5. Conclusions

In the image processing the common DCT blocks of 8x8 are considered. As images, in general, may not be sparse, two approaches are used. One is to make the analysed image to be sparse in the 2D-DCT domain, by considering only a few of its largest values as non-zero. The processing and reconstruction is then based on this picture. Since common images contain just a few significant 2D-DCT coefficients in the 8x8 blocks this step can be also avoided and the original image, without pre-processing can considered as sparse. This is done in the second approach presented in this paper. The reconstruction was done with compressive sensing gradient algorithm. Both approaches produced similar results for the analysed image. The case of missing samples can be considered due to reduced number of available samples or due to our desire to use a small number of signal samples in the process of information storage or transmission. One more case that reduces to the compressive sensing formulation is the presence of strong, salt-and-paper, noise in the images. After the noisy pixels are detected they can be discarded and considered as missing. The compressive algorithm is then used to recover the image original pixels. It has been shown that the recovery can be succesful with up to 75% missing pixels. The algorithm has been applied on both black and white and colour images. In both cases the images are recovered with a high accuracy.

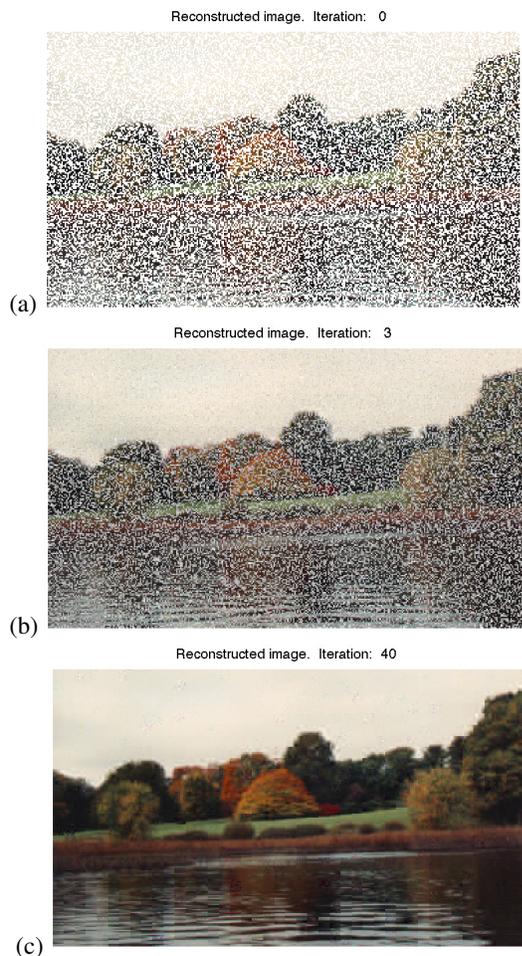

(a)

(b)

(c)

Figure 12: (a) Image with missing/corrupted samples; (b) Reconstruction after 3 iterations; (c) Reconstructed image